\documentclass[letterpaper]{article} 
\usepackage{aaai23}  
\usepackage{times}  
\usepackage{helvet}  
\usepackage{courier}  
\usepackage[hyphens]{url}  
\usepackage{graphicx} 
\usepackage{natbib}  
\usepackage{caption} 
\usepackage{algorithm}
\usepackage{algorithmic}

\usepackage{newfloat}
\usepackage{listings}
\DeclareCaptionStyle{ruled}{labelfont=normalfont,labelsep=colon,strut=off} 
\floatstyle{ruled}
\newfloat{listing}{tb}{lst}{}
\floatname{listing}{Listing}
\usepackage{color}

\usepackage[hidelinks]{hyperref}
\usepackage{amsmath}
\usepackage{amsfonts}
\usepackage{amssymb}

\title{NCTV: Neural Clamping Toolkit and Visualization\\ for Neural Network Calibration}
\author{
    Lei Hsiung\textsuperscript{\rm 1,3},
    Yung-Chen Tang\textsuperscript{\rm 1,2},
    Pin-Yu Chen\textsuperscript{\rm 3},
    Tsung-Yi Ho\textsuperscript{\rm 1,4}
}
\affiliations{
    \textsuperscript{\rm 1}National Tsing Hua University \\
    \textsuperscript{\rm 2}MediaTek Inc. \\
    \textsuperscript{\rm 3}IBM Research \\
    \textsuperscript{\rm 4}The Chinese University of Hong Kong \\
    \{hsiung, yctang\}@m109.nthu.edu.tw, pin-yu.chen@ibm.com, tyho@cse.cuhk.edu.hk
}

\begin{document}

\maketitle
\begin{abstract}
With the advancement of deep learning technology, neural networks have demonstrated their excellent ability to provide accurate predictions in many tasks. However, a lack of consideration for \textit{neural network calibration} will not gain trust from humans, even for high-accuracy models. In this regard, the gap between the confidence of the model's predictions and the actual correctness likelihood must be bridged to derive a well-calibrated model. In this paper, we introduce the Neural Clamping Toolkit, the first open-source framework designed to help developers employ state-of-the-art model-agnostic calibrated models. Furthermore, we provide animations and interactive sections in the demonstration to familiarize researchers with calibration in neural networks. A Colab tutorial on utilizing our toolkit is also introduced.
\end{abstract}

\section{Introduction}\label{sec:introduction}
With the increasing number of tasks that deep neural networks can handle, including medical diagnosis, image classification, natural language processing, etc., it is indispensable to increase the human trustworthiness of AI models. For example, if an AI model classifies a pathological image as malignant, a radiologist might need to know what it is based on and how likely that the prediction is correct. A confidence level is, therefore, an essential basis for physicians and radiologists to perform disease diagnosis or tumor analysis on medical images \cite{jiang2012calibrating, esteva2017dermatologist}. Because the risk will be magnified in safety-related scenarios, it is also crucial to provide accurate prediction confidence.

However, most neural networks are not required to possess accurate confidence values when trained or landing, resulting in the potential risk of bias between confidence values and accuracy. To address this concern, one of the approaches is through \textit{neural network calibration}, that is, making the confidence of model prediction align with its true correctness likelihood \cite{guo2017calibration}. 

In the existing literature, two calibration approaches have been proposed: the \textit{in-processing} way involves training or fine-tuning the model \cite{tian2021geometric, liang2020improved}, and the \textit{post-processing} way mainly focuses on processing or remapping the output of the pre-trained model logits \cite{guo2017calibration, esteva2017dermatologist, kull2019beyond, gupta2020calibration}. However, these methods are either time-consuming and computationally expensive or suffer from a lack of effectiveness. Therefore, \cite{tang2022neuralclamping} proposed \textit{Neural Clamping}, a novel post-processing calibration framework for neural networks, the first approach that utilizes a joint input-output transformation for model-agnostic calibration. At the model input and output, respectively, Neural Clamping appends a universal perturbation and temperature scaling for all classes. It provides a novel framework for post-processing calibration and includes temperature scaling as a special case. In addition, they also develop theoretical analyses and experiments to prove that this method can effectively reduce calibration error.

\begin{figure}[t]
    \centering
    \includegraphics[width=\linewidth]{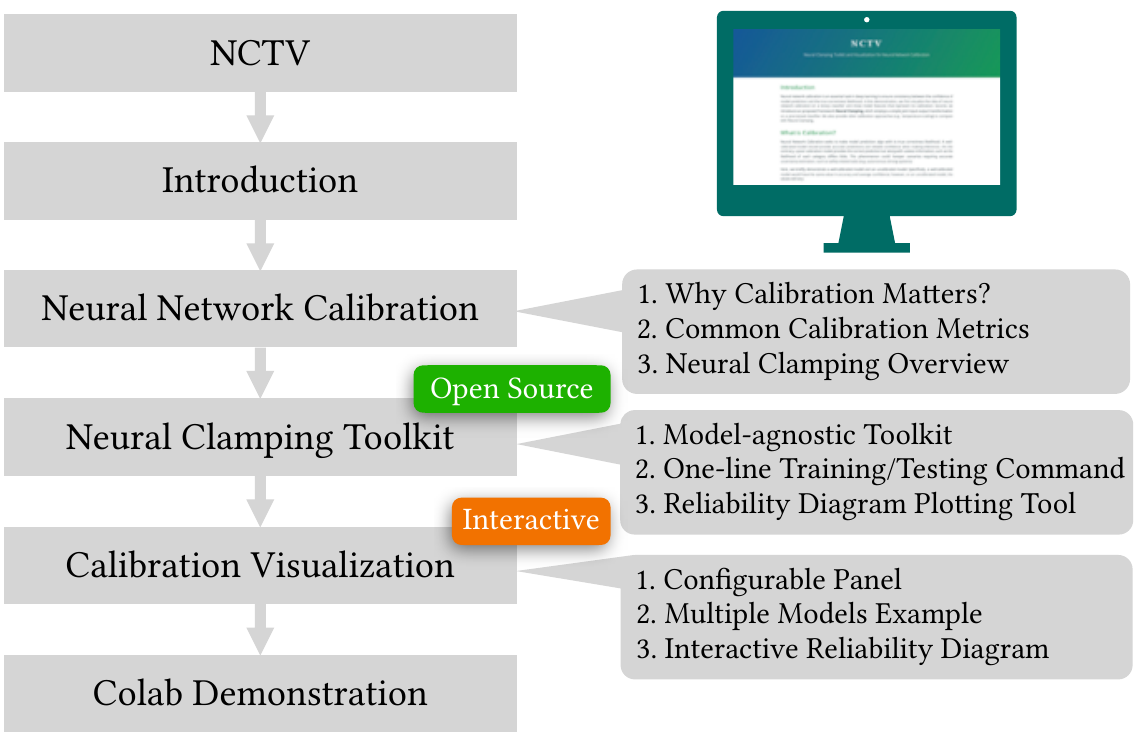}
    \caption{NCTV Overview. Browse on: \href{https://hsiung.cc/NCTV/}{hsiung.cc/NCTV}}
    \label{fig:demonstration-overview}
\end{figure}

Extending on this idea, in this paper, we present \textit{NCTV: Neural Clamping Toolkit and Visualization for Neural Network Calibration}, the first open-source framework and web-based demonstration to familiarize researchers and users with neural network calibration. As shown in Figure \ref{fig:demonstration-overview}, NCTV also includes an interactive section that allows the user to observe real-time changes in the reliability diagram by manipulating the calibration tool in the configurable panel. We have also created a step-by-step notebook tutorial on Google Colab to guide users using our toolkit on their own models and visualize the performance before and after calibration in reliability diagrams.

\section{NCTV: The Overview}\label{sec:nctv}
\subsection{Neural Network Calibration}
\begin{figure}[t]
    \centering
    \includegraphics[width=\linewidth]{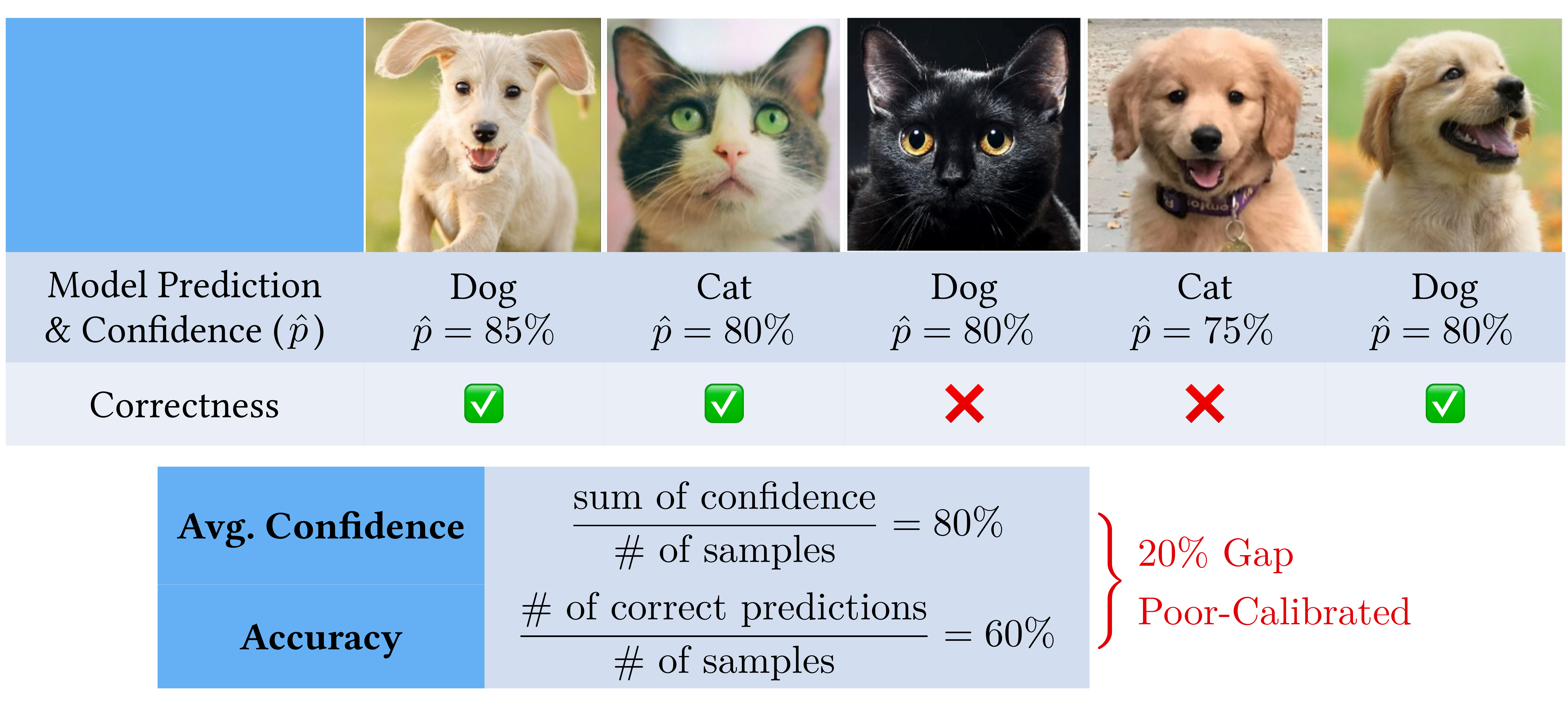}
    \caption{The example of poor-calibrated neural network.}
    \label{fig:poor_calibrated}
\end{figure}

Given a $K$-way neural network classifier $f_{\theta}: \mathcal{X} \to \mathbb{R}^{K}$, and $x\in\mathcal{X}$ is an data sample with ground-truth $y\in\{0,\cdots,K\}$. Let $\hat{y}$ and $\hat{p}$ denote the model $f_{\theta}$ predicts the most likely class of $x$ and its confidence, we called that $f_{\theta}$ is \textit{calibrated} if $\mathbb{P}(y=\hat{y}|p=\hat{p})=\hat{p}$. Therefore, a poor-calibrated neural network might have poor alignment between the model's predictions and its confidence levels. As shown in Figure \ref{fig:poor_calibrated}, the model is poorly calibrated because there is a wide gap between its accuracy and average confidence level.

\subsection{Neural Clamping Toolkit (\texttt{NCToolkit})}
\begin{figure}[t]
    \centering
    \includegraphics[width=\linewidth]{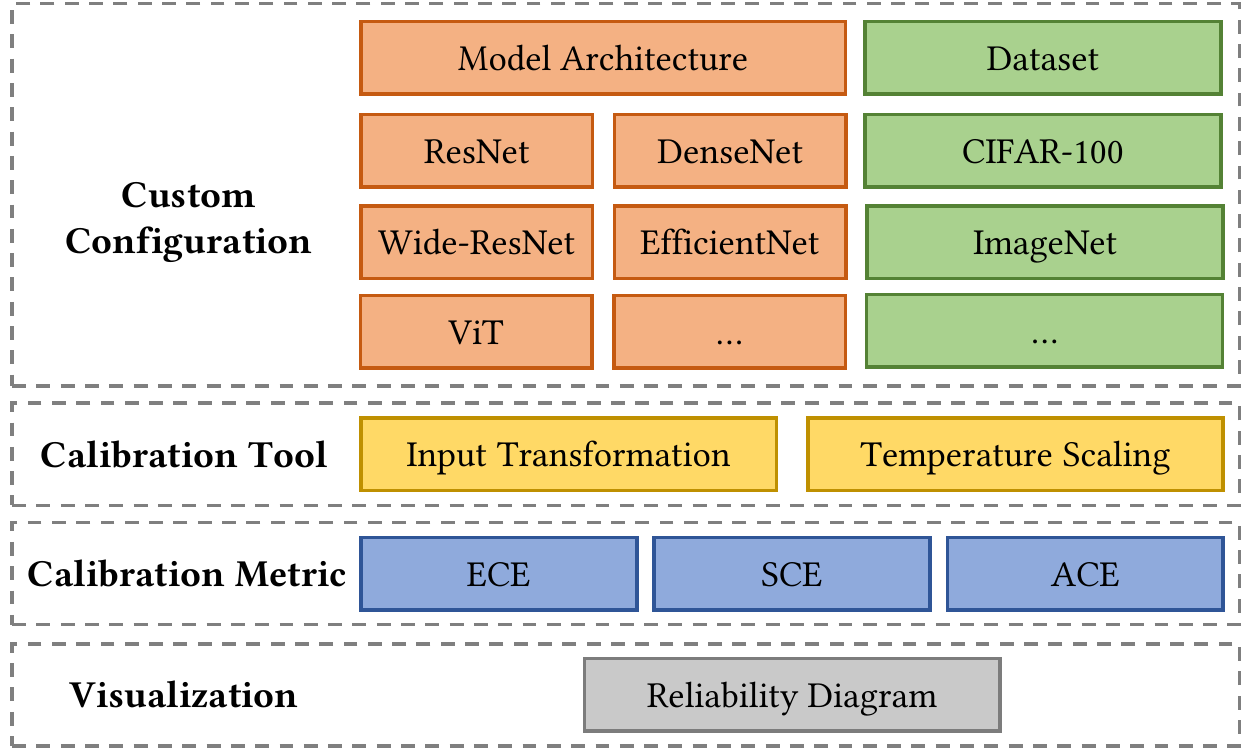}
    \caption{\texttt{NCToolkit} overall framework.}
    \label{fig:NCToolkit}
\end{figure}

\texttt{NCToolkit} possesses three characteristics: 1) it is a model-agnostic framework so that it can be directly applied to any pre-trained model, 2) it is developed in a highly modularized framework, so it is convenient for researchers to extend and operate, and 3) it could significantly outperform state-of-the-art post-processing calibration methods \cite{tang2022neuralclamping}.
The overall framework of \texttt{NCToolkit} is shown in Figure \ref{fig:NCToolkit}, which is composed of four main components. We introduce each component in the following.

\paragraph{Custom Configuration:} 
\texttt{NCToolkit} is developed in the PyTorch framework, therefore, we support common model architectures and datasets. We provide the default configuration, including our pre-trained models and pre-trained Neural Clamping parameters for CIFAR-10 and ImageNet Datasets. Users can modify or extend upon specific needs.

\paragraph{Calibration Tool:}
We combine input perturbation and temperature scaling for Neural Clamping. This integration could offer state-of-the-art post-processing calibration tool. In \texttt{NCToolkit}, user could simply call \texttt{.train\_NC()} to start calibrating.

\paragraph{Calibration Metric:}
To properly measure the model calibration, the most common metric is the Expected Calibration Error (ECE), which could be defined as $\mathbb{E}_{(x,y)\sim\mathcal{D}}(|\mathbb{P}(y=\hat{y}|p=\hat{p})-\hat{p}|)$. Other popular metrics, such as Static Calibration Error (SCE) and Adaptive Calibration Error (ACE), are also available for developers to evaluate calibration performance.

\definecolor{pink}{rgb}{1.0, 0.6437, 0.7086}
\definecolor{blue}{rgb}{0.2196, 0.2196, 1.0}

\begin{figure}[t]
    \centering
    \includegraphics[width=\linewidth]{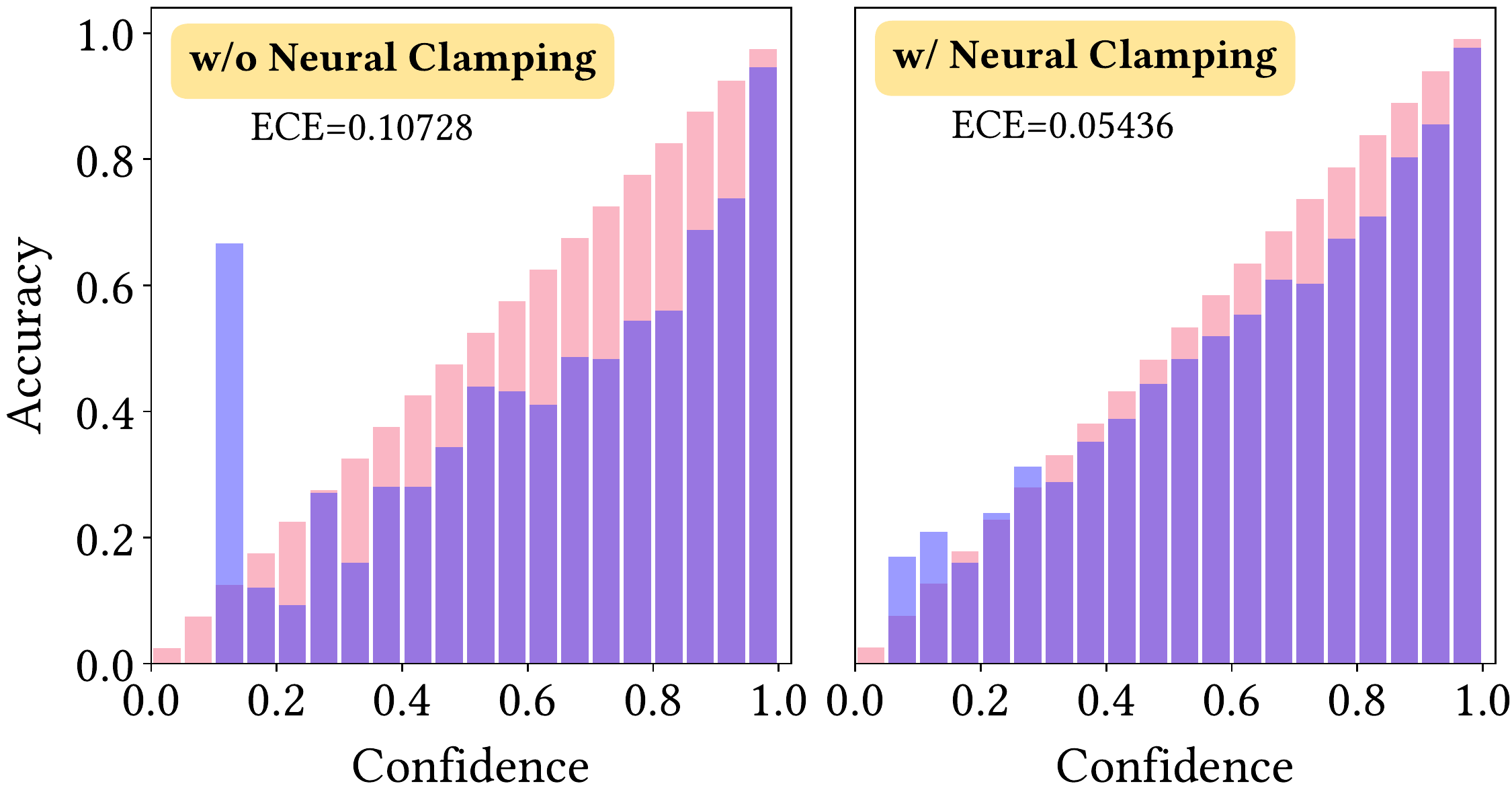}
    \caption{Reliability diagram examples of a neural network with and without applied Neural Clamping for calibration. The bar marked in  \textcolor{blue}{blue} (\textcolor{pink}{pink}) denotes the actual (expected) sample accuracy, which is a function of confidence. Deviations from the \textcolor{pink}{expected bar} line indicate calibration errors.}
    \label{fig:reliability_diagram}
\end{figure}

\paragraph{Visualization:}
We provide a visualization tool to depict calibration performance in \textit{reliability diagrams} \cite{degroot1983comparison, niculescu2005predicting}. As shown in Figure \ref{fig:reliability_diagram}, the reliability diagram is composed of $M$ bins, where each bin $B_m$ represents the set of samples, whose prediction confidence falls within the interval $I_m = (\frac{m- 1}{M },\frac{m}{M}]$. Specifically, the accuracy of each bin can be defined as: 
\begin{equation*}
    \text{acc}(B_m) = \frac{1}{|B_m|}\sum_{i\in{B_m}}{\mathbf{1}(\hat{y_i}=y_i)},
\end{equation*}
where $|B_m|$ represents the number of samples in $B_m$, and $\hat{y_i}$ and $y_i$ are the most likely and ground-truth labels for sample $i\in{B_m}$. Accordingly, the diagram will vary with the number of bins, and so does the corresponding ECE.

\section{Conclusion}\label{sec:conclusion}
This demonstration enables users to gain a deeper understanding of neural network calibration.
\texttt{NCToolkit} features a modularized and extensible framework to support developers in calibrating their model.
In our demonstration, we also provide Colab tutorials on using our toolkit. Furthermore, the visualization tool is also introduced to allow users to depict the calibration performance of the model.

\section*{Acknowledgments}
Part of this work was done during Lei Hsiung's visit to IBM Thomas J. Watson Research Center.

\bibliography{aaai23}

\end{document}